\documentclass[10pt,twocolumn,letterpaper]{article}

\usepackage{cvpr}
\usepackage{times}
\usepackage{graphicx}
\usepackage{amsmath}
\usepackage{amssymb}
\usepackage[vlined,ruled,linesnumbered]{algorithm2e}

\newcommand{\pname}[1]{{\textsc{REC}}{#1}}
\cvprfinalcopy
\begin{document}

\title{Regularize, Expand and Compress: Multi-task based Lifelong Learning via NonExpansive AutoML}

\author{Jie Zhang$^1$, Junting Zhang$^2$, Shalini Ghosh$^3$, Dawei Li$^3$, Jingwen Zhu$^3$, Heming Zhang$^2$, Yalin Wang$^1$\\
$^1$Arizona State University, Tempe, AZ; $^2$University of Southern California, Los Angeles, CA;\\ $^3$ Samsung Research America, Mountain View, CA}
\maketitle

\begin{abstract}
Lifelong learning, the problem of continual learning where tasks arrive in sequence, has been lately attracting more attention in the computer vision community. The aim of lifelong learning is to develop a system that can learn new tasks while maintaining the performance on the previously learned tasks. However, there are two obstacles for lifelong learning of deep neural networks: catastrophic forgetting and capacity limitation. To solve the above issues, inspired by the recent breakthroughs in automatically learning good neural network architectures, we develop a Multi-task based lifelong learning via nonexpansive AutoML framework termed Regularize, Expand and Compress (REC). REC is composed of three stages: 1) continually learns the sequential tasks without the learned tasks' data via a newly proposed multi-task weight consolidation (MWC) algorithm; 2) expands the network to help the lifelong learning with potentially improved model capability and performance by network-transformation based AutoML; 3) compresses the expanded model after learning every new task to maintain model efficiency and performance. The proposed MWC and \pname{} algorithms achieve superior performance over other lifelong learning algorithms on four different datasets. 
\end{abstract}

\section{Introduction}
In many real-world applications, batches of data arrive periodically (e.g., daily, weekly, or monthly) with the data distribution changing over time. This presents an opportunity for lifelong learning or continual learning, and is an important developing topic of interest in artificial intelligence. The primary goal of lifelong learning is to learn consecutive tasks without forgetting the knowledge learned from previously trained tasks, and leverage the previous knowledge to obtain better performance or faster convergence on the newly coming task. One simple way is to fine-tune the model for every new task; however, such retraining typically degenerates the model performance on both new tasks and the old ones. If the new tasks are largely different from the old ones, it might not be able to learn the optimal model for the new tasks. Meanwhile, the retrained representations may adversely affect the old tasks, causing them to drift from their optimal solution.
This can cause ``catastrophic forgetting", a phenomenon where training a model with new tasks interferes the previously learned old knowledge, leading to a performance degradation or even overwriting of the old knowledge by the new one. 

\begin{figure}[t]
\centering
\includegraphics[width=8cm]{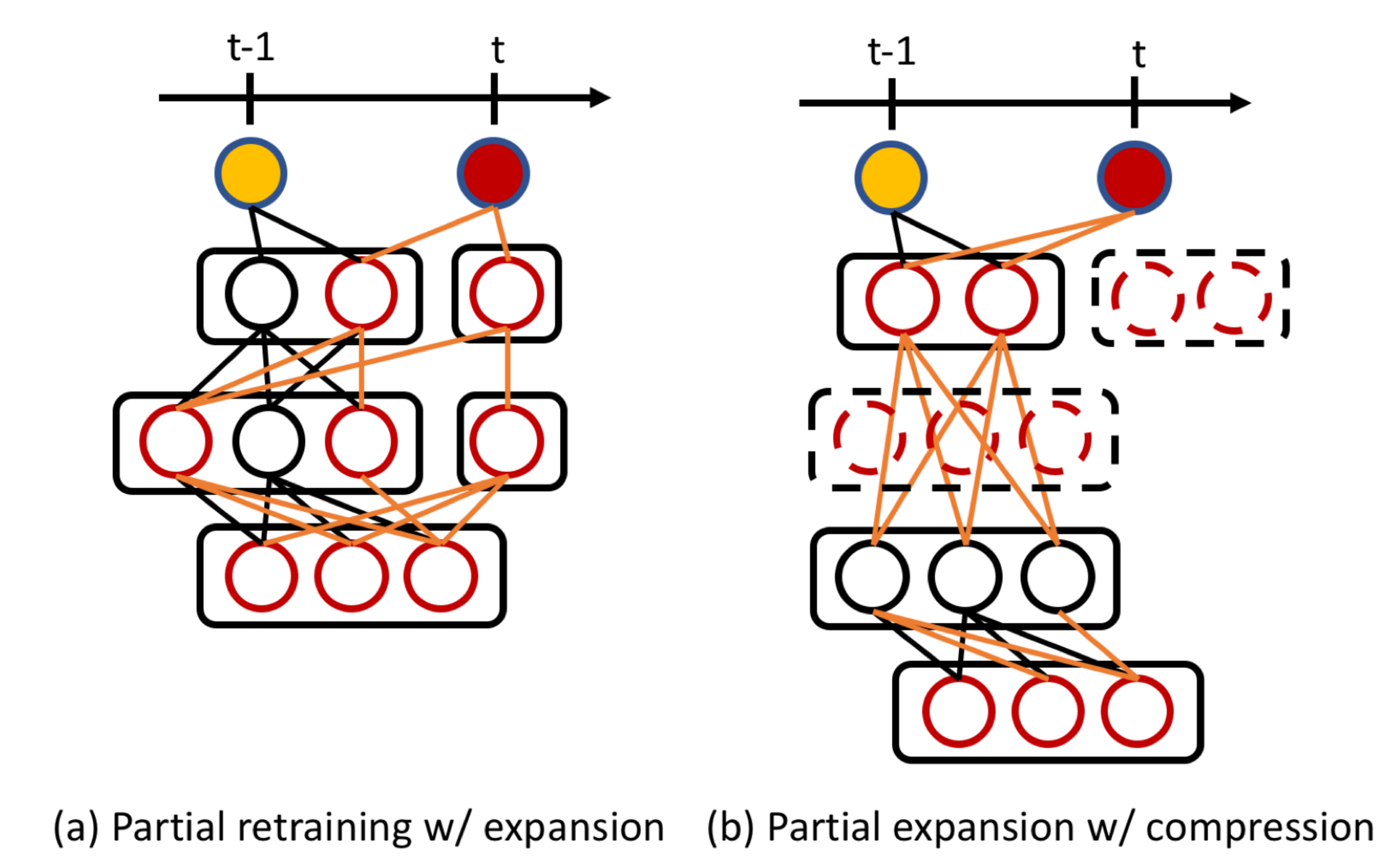}
\caption{(a) State-of-the-art DEN method~\cite{yoon2018lifelong} selectively retrains the old network, dynamically expands the model capacity (b) The proposed \pname{} method expands the network through network transformation based AutoML, and then subsequently compresses the model to its original size.}
\label{fig:1}
\vspace{-1em}
\end{figure}

To overcome above catastrophic forgetting problem, many approaches have been proposed ~\cite{kirkpatrick2017overcoming,li2017learning,rebuffi2017icarl}. Kirkpatrick \etal \cite{kirkpatrick2017overcoming} propose using a regularization term to prevent the new weights from deviating too much from the previously learned weights, based on their significance to old tasks. Their method uses a fixed neural network architecture, which would not scale up when network capacity gets saturated with more and more new tasks to learn. Dynamically expanding the network~\cite{yoon2018lifelong} (DEN) is one way to overcome the problem caused by static architecture --- it expands the network capacity whenever it detects that the loss for the new task would not reach a pre-defined threshold. However, DEN involves many hyperparameters and the final performance is highly sensitive to these parameters; it relies on hand-crafted heuristics to explore the tuning space. But the search space is considerably large, such that human experts usually find a sub-optimal solution while the current parameters tuning procedures are time-consuming. To this end, we aim to automatically expand the network for lifelong learning, with higher performance and less parameter redundancy than human-designed architectures. To better facilitate (a) automatic knowledge transfer without human expert tuning and (b) model design with optimized model complexity, we unprecedentedly propose to apply AutoML~\cite{robert2014machine} for lifelong learning while taking learning efficiency into consideration.

AutoML refers to automatically learn a suitable machine learning (ML) model for a given task --- Neural Architecture Search (NAS)~\cite{zoph2016neural} is a subfield of AutoML for deep learning, which searches for optimal hyperparameters of designing a network architecture using reinforcement learning (RL). The RL framework has a main controller that observes the generated children networks' performance on the validation set as the reward signal, it then gives higher probabilities to architectures that have higher performance than the lower ones to update the model. If we use this approach directly in the lifelong learning setting, it would forget old tasks' knowledge and be a wasteful process since each new task network architecture would need to be searched from scratch by the controller, ignoring the correlations between previously learned tasks and the new task. We hereby propose a multi-task weight consolidation (MWC) approach to learn the discriminative weights subset by incorporating inherent correlations between old tasks and new task. Furthermore, to narrow down the architecture searching space and save training time, network transformation based AutoML~\cite{cai2018efficient} is utilized to accelerate the meta-learning of the new network. 

However, if we keep expanding the network for more and more new tasks, the model will have a much larger model size comparing with the initial model and suffer the inefficient problem (e.g., low memory footprint, low power usage). Many network-expanding-based lifelong learning algorithms~\cite{rusu2016progressive, yoon2018lifelong} increase the model capability but also decrease the learning efficiency in terms of memory cost and power usage. To address this issue, we conduct model compression after completing the learning of each new task --- we compress the expanded model to the initial model, with negligible performance loss on both old and new tasks. Fig~\ref{fig:1} shows the main difference of our approach with network expansion-based lifelong learning algorithms.

In this paper, we propose a Multi-task based lifelong learning via nonexpansive AutoML framework termed Regularize, Expand and Compress (REC), to continually and automatically learn on such sequential data sets. We start with a given small network to learn an initial model on the first given task; \pname{} then searches the best network architecture by network transformation based AutoML for the new upcoming task without access to the old tasks' data using a newly proposed MWC algorithm and compress the expanded network size to the initial network size. 

Our key contributions of this work can be summarized as follows:
\begin{itemize}
\item We propose to Regularize, Expand and Compress (\pname{}) for lifelong learning, which automatically expands the network capacity for learning a new task with higher performance and less parameter redundancy than human-designed architectures. 
\item To overcome catastrophic forgetting for the old learned tasks, we propose a novel Multi-task Weight Consolidation (MWC) --- it considers the discriminative weight subset by incorporating inherent correlations between old tasks and new task and learns the newly added layer as a task-specific layer for the new task.
\item Furthermore, unlike previous network-expanding-based lifelong learning algorithms, \pname{} compresses the model after learning every new task to guarantee the model efficiency. The final model is a non-expensive model but the performance enhanced by network expanding before the compression. 
\end{itemize}


 \begin{table*}[t]
\centering
\caption{Comparisons of the lifelong learning approaches for overcoming catastrophic forgetting. EWC: Elastic Weight Consolidation~\cite{kirkpatrick2017overcoming}; DEN: Dynamically expandable network~\cite{yoon2018lifelong}; LwF: Learning without forgetting~\cite{li2017learning}; GEM: Gradient of Episodic Memory~\cite{lopez2017gradient}; PGN: Progressive neural network~\cite{rusu2016progressive} and our algorithm \pname{}. }
\label{table:comp-lifelong}
 \begin{tabular}{c|cccccc}\hline
	& EWC & DEN & LwF & GEM  & PGN & \pname{} \\ \hline
No memory growth &\checkmark&&\checkmark&\checkmark&&\checkmark\\
No exemplar & \checkmark & \checkmark & \checkmark &  & \checkmark&\checkmark\\
Expanding network capacity when necessary& & \checkmark &  & & \checkmark&\checkmark \\
AutoML ability &&&&&&\checkmark\\\hline
\end{tabular}
\end{table*}

\section{Related Work}
\subsection{Overcoming Catastrophic Forgetting}
Recently, a lot of lifelong learning methods were proposed to address the catastrophic forgetting problem. The first group of methods uses regularized learning. Elastic Weight Consolidation (EWC)~\cite{kirkpatrick2017overcoming} shows that task-specific synaptic consolidation may overcome catastrophic forgetting in neural networks and observes the important weights for the previous tasks and selectively adjusts the plasticity of the weights. Inspired by EWC, Schwarz \etal~\cite{schwarz2018progress} propose online EWC, which enlarges the EWC scalability by limiting the regularization term computational cost when the number of tasks increases. Synaptic Intelligence~\cite{zenke2017continual} computes an online importance measure along an entire learning trajectory, which is similar to EWC. Rotate-EWC~\cite{liu2018rotate} (REWC) is a modified version of EWC --- it approximately diagonalizes the Fisher information matrix of the network parameters that compute the factorized rotation of the parameter space used in conjunction with EWC. 

The second group of the strategies is associated with learning task-specific parameters. Learning without forgetting (LwF)~\cite{li2017learning} leverages distillation regularization on the new tasks --- the soft labels of previously learned tasks are enforced to be similar to the network with the current task by using knowledge distillation~\cite{hinton2015distilling}. Less-forgetful learning~\cite{jung2017less} is proposed to regularize the $L_2$ distance between the final hidden activations and the old tasks' parameters for preserving the old task feature mappings. 

The third group of methods expands the network capacity. Progressive neural network (PGN)~\cite{rusu2016progressive} is proposed to block any changes to the pre-trained network models on previously learned tasks and expands the network architecture by allocating sub-networks with the fixed capacity to be trained with the new information. PathNet~\cite{fernando2017pathnet} uses agents embedding into a neural network to find which parts of the network can be reused for learning new tasks and freezes task-relevant paths for avoiding catastrophic forgetting. Dynamically expanding network (DEN)~\cite{yoon2018lifelong} increases the number of trainable parameters to continually learn new tasks and dynamically selects neurons to retrain or expand neuron capacity by using group sparse regularization. 

The other family of the methods uses episodic memory, where the previously learned task samples are stored to effectively recall the experience in the past. Gradient of Episodic Memory (GEM)~\cite{lopez2017gradient} performs positive forward transfer, minimizes negative backward transfer to previously learned tasks and learns the subset of correlations to a set of tasks without using task descriptors. Incremental Classifier and Representation Learning (iCaRL)~\cite{rebuffi2017icarl} combines classification loss on new tasks and distillation loss on previously learned tasks with a K-nearest neighbor classifier and selects the exemplars for each task by letting the embeddings of the selected samples closer to the center point of each class. Table~\ref{table:comp-lifelong} shows the multiple merits of \pname{}, comparing with previous researches in this area. 

\subsection{AutoML and Knowledge Distillation}
There are many works on AutoML to improve the performance of deep neural networks~\cite{zoph2016neural,pham2018efficient,cai2018efficient}. Neural Architecture Search (NAS)~\cite{zoph2016neural} searches the transferable network blocks via reinforcement learning and outperforms many manually designed network architecture. ENAS~\cite{pham2018efficient} uses a controller to discover network architectures by searching an optimal subgraph within a large computational graph and shares parameters among child models to enable efficient NAS. EAS~\cite{cai2018efficient} efficiently explores network architecture via network transformation~\cite{chen2015net2net} which is a functionality preserving method to expand the architecture with a fixed number of units or filters. 

Besides, Knowledge distillation (KD)~\cite{hinton2015distilling} is also very related to our work. KD is widely used to compress a network with a different architecture that approximates the original network where knowledge is transferred from a large teacher network to a small student network. The student network is trained with KD loss --a modified cross-entropy loss-- that ensures the teacher network and student network are similar. In our work, we adopt the KD to compress the expanded network after learning each new task.  

\section{Method}
\begin{figure}[t]
\centering
\includegraphics[width=8cm]{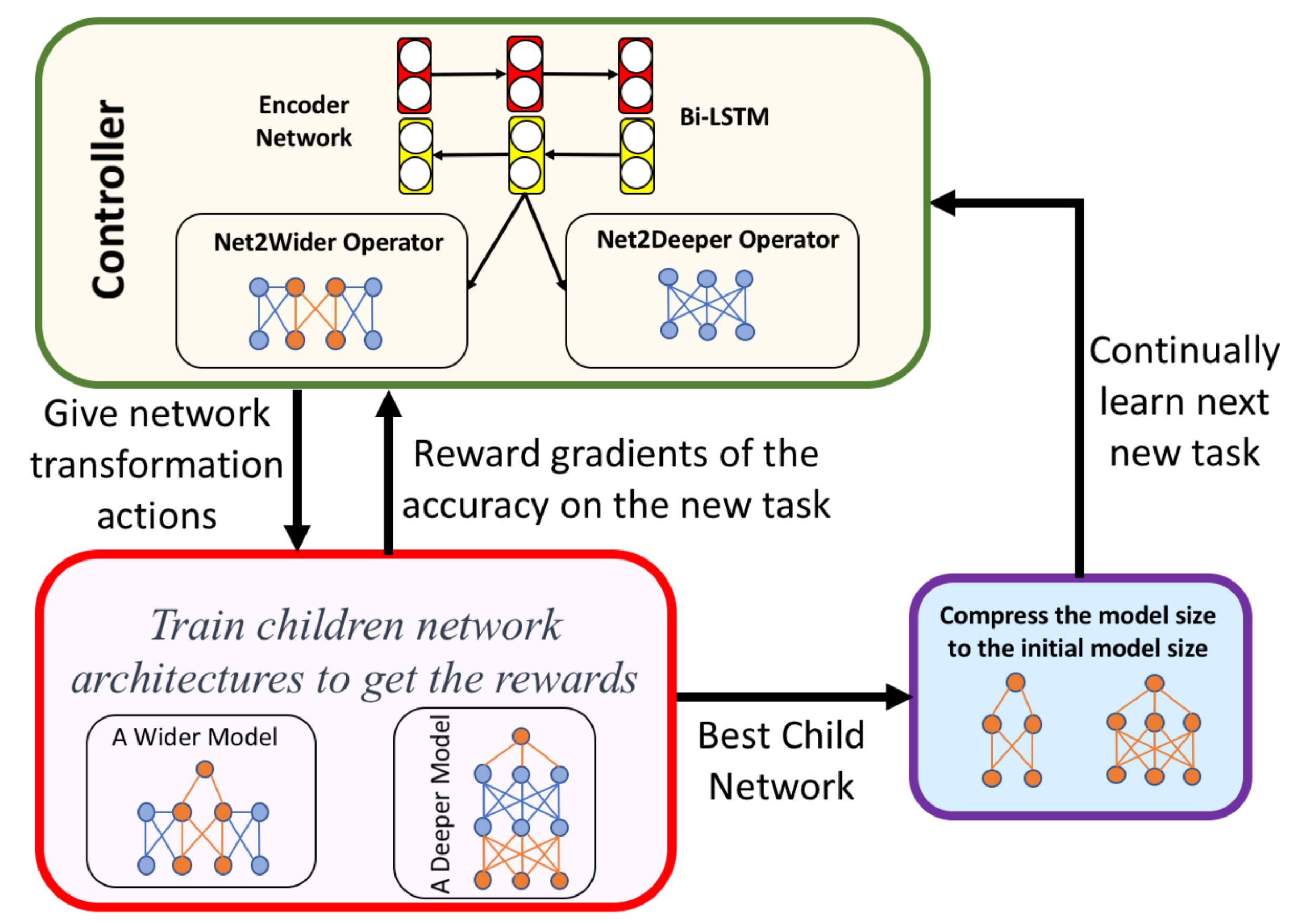}
\caption{Illustration of our lifelong learning framework. \pname{} first uses MWC to search the best child network by Net2Deeper and Net2Wider operators in the controller for a new coming task, then compresses the expanded network to the same size as the initial model and continually learns next new task.}
\label{fig:lifecam}
\end{figure}

Fig.~\ref{fig:lifecam} is an overview of our AutoML framework \pname{} for lifelong learning, it has three steps: Regularize multi-task weight consolidation, Expand network by AutoML and Compress the expanded model. 

\subsection{Problem Definition and Overview}
We define the lifelong learning problem as follows --- there will be an unknown number of tasks with unknown distributions, arriving in sequence. Our goal is to learn a deep model in such a lifelong learning scenario without catastrophic forgetting. For the evaluation protocol, we report the classification accuracy of each of previous $T-1$ tasks and the current task $T$ after training on the $T$-th task. Given a sequence of $T$ tasks, task at time point $t = 1, 2, \cdots, T$ with $N_t$ images comes with dataset $\mathbf{D}_t = \{\mathbf{x}_i^t, y_i^t\}_{i=1}^{N_t}$. Specifically, for task $t$, $y_i^t \in \{1, ..., K\} $ is the label for the $i$-th sample $\mathbf{x}_i^t \in \mathbb{R}^{d_t}$ in task $t$. We denote the training data matrix by $\mathbf{X}^t$ for $\mathbf{D}_t$, i.e., $\mathbf{X}^t = (\mathbf{x}^t_1, \cdots, \mathbf{x}^t_{N_t}).$ When the dataset of task $t$ comes, all the previous training datasets $\mathbf{D}_1, \cdots, \mathbf{D}_{t-1}$ are not available any more, but the deep model parameter $\theta^{t-1} = \{\theta_l^{t-1}\}_{l=1}^L$  can be accessed. The lifelong learning problem at time point $t$ when given data $\mathbf{D}_t$ can be defined as solving the following problem:
\begin{equation}
    \min_{\theta_t} {\cal F}(\theta_t |\theta_{t - 1} , \mathbf{D}_t), t = 1, \cdots, T
    \label{eq:lifelong-obj}
\end{equation}
where $\mathcal{F}$ is the loss function of solving $\theta^t$, $\theta^t$ is the parameter for task $t$. Note that the number of the upcoming tasks can be finite or infinite --- for simplification, we consider the finite scenario here. 

Kirkpatrick et al.~\cite{kirkpatrick2017overcoming} proposed EWC that consists of a quadratic penalty on the difference between the parameter $\theta^t$ and $\theta^{t-1}$ to slow down the catastrophic forgetting for previously learned tasks. The posterior distribution $p(\theta^t|\mathbf{D}_t)$ is used to describe the problem by the Bayes' rule.
\begin{equation}
    \log p(\theta^t |\mathbf{D}_t) =  \log p(\mathbf{D}_t |\theta^t) + \log p(\theta^t | \mathbf{D}_{t-1}) - \log p(\mathbf{D}_t),
\label{eq:EWC-bayes}
\end{equation}
where the posterior probability $\log p(\theta^t | \mathbf{D}_{t - 1})$ embeds all the information from task $t-1$. However, the problem~(\ref{eq:EWC-bayes}) is intractable so that EWC approximates it as a Gaussian distribution with mean of parameter $\bar{\theta}^{t-1}$ and a diagonal $I$ of the Fisher Information matrix $\mathbb{F}$. The matrix $\mathbb{F}$ is computed by $\mathbb{F}_i = I(\theta^t)_{ii} = E_x[(\frac{\partial}{\partial \theta^{t}_i}\log p(\mathbf{D}_t|\theta^t))^2|\theta^t]$. Therefore, the problem of EWC on task $t$ can be written as follows:
\begin{equation}
    \min_{\theta^t} \quad \mathcal{F}_t(\theta^t) 
    + \frac{\lambda}{2}\sum_{i}\mathbb{F}_{i}(\theta^t_{i} - \bar{\theta}^{t-1}_i)^2,
\label{eq:EWC}
\end{equation}
where ${\cal F}_t$ is the loss function for task $t$, $\lambda$ denotes how important the task $t-1$ is compared to the task $t$ and $i$ labels each weight of the parameter $\theta$.

\begin{figure}[t]
\centering
\includegraphics[width=8cm]{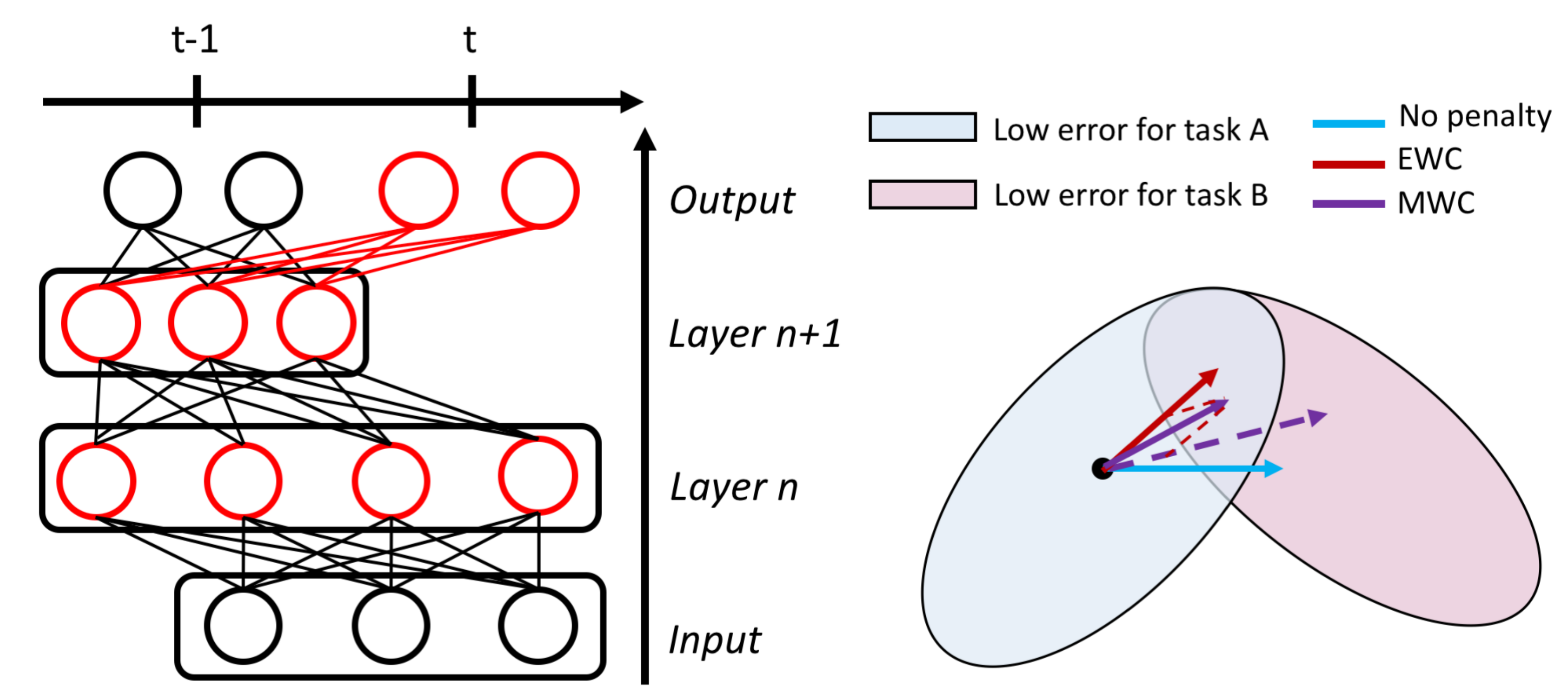}
\caption{MWC retrains the entire network learned on previous tasks while regularizing it to prevent forgetting from the original model. MWC (purple solid line) learns better parameter representations to overcome catastrophic forgetting by studying MTL with the sparsity-inducing norm (purple dash line) and EWC (red line).}
\label{fig:4}

\end{figure}

\subsection{Multi-task Weight Consolidation}
The main problem of EWC is that EWC only enforces task $t$ close to task $t-1$. This will ignore the inherent correlations between task $t-1$ and task $t$ and such relationship might potentially help overcome catastrophic forgetting on the previously learned tasks. Learning multiple related tasks jointly can improve performance relative to learning each task separately, when the tasks are related --- this idea is incorporated into Multi-Task Learning (MTL)~\cite{evgeniou2004regularized}. It has been commonly used to obtain better generalization performance than learning each task individually. We redefine Eq.~\ref{eq:EWC} using MTL and propose a new objective function Eq.~\ref{eq:EWC-MT} to improve the ability of overcoming catastrophic forgetting from multiple tasks simultaneously: 
\begin{equation}
\begin{aligned}
    \min_{\theta^t} \mathcal{F}_t(\theta^t) 
    + \frac{\lambda}{2}\sum_{i}\mathbb{F}_{i}(\theta^{t}_i - \bar{\theta}^{t-1}_i)^2 +  \lambda_2||[\theta^{t}; \theta^{t-1}]||_{2, 1},
    \end{aligned}
    \label{eq:EWC-MT}
\end{equation}
where $\lambda_2$ is the non-negative regularization parameter and $||[\theta^t; \theta^{t-1}]||_{2, 1}=||||\theta^t||_2,||\theta^{t-1}||_2||_1$  is the $l_{2, 1}$-norm regularization to learn the related representations. Here, we employ the multi-task learning with $l_{2, 1}$-norm~\cite{liu2009multi} to capture the common subset of relevant parameters from each layer for task $t-1$ and task $t$. 

Specifically, we further consider some important parameters which have better representation power to a subset of tasks. The MTL with sparsity-inducing norm~\cite{gong2012multi} has been widely studied to select such discriminative parameter subset by incorporating inherent correlations among multiple tasks. To this end, the $l_1$ sparse norm is imposed to learn the new task-specific parameters while learning task relatedness among multiple tasks. Therefore, the objective function for task $t$ becomes:
\begin{equation}
\begin{aligned}
    \min_{\theta^t} \quad &\mathcal{F}_t(\theta^t) 
    + \frac{\lambda}{2}\sum_{i}\mathbb{F}_{i}(\theta^{t}_i - \bar{\theta}^{t-1}_i)^2 \\+&  \lambda_2||[\theta^{t};\theta^{t-1}]||_{2,1}+\lambda_3||\theta^t||_1  ,
    \end{aligned}
    \label{eq:MWC}
\end{equation}
where $\lambda_3$ is the non-negative regularization parameter. We call our algorithm Multi-task Weight Consolidation (MWC) because it studies the discriminative \textit{weights} subset with inherent correlations among \textit{multiple tasks}. Fig.~\ref{fig:4} shows the geometric illustration of MWC. 

\subsection{AutoML for Lifelong Learning with MWC}
\begin{algorithm}[t]
\caption{\pname{}}
\label{alg:1}
\SetKwData{Left}{left}\SetKwData{This}{this}\SetKwData{Up}{up}
\SetKwFunction{Union}{Union}\SetKwFunction{FindCompress}{FindCompress}
\SetKwInOut{Input}{Input}\SetKwInOut{Output}{Output}
\Input{Dataset $\mathbf{D}_1, \cdots, \mathbf{D}_T, \lambda, \lambda_1, \lambda_2$ }
\Output{ $\theta^{T}_c$}
\Begin{
   \For{ $t=1 \rightarrow T$ }{
   \If{ $t=1$ }{
       Train an initial network with weights $\theta^1$ by using Eq.~\ref{eq:lifelong-obj}.\\
       }
    \Else{
        Search a best child network $\theta^t$ by Alg.~\ref{alg:Net2Net} with Eq.~\ref{eq:modified-MWC}.\\
        Compress $\theta^t$ to the same model size as $\theta^1$ by Eq.~\ref{eq:kd} and use $\theta^t_c$ for next task.\\
      }
  }
}
\end{algorithm}
MWC is a regularization-based lifelong learning algorithm, it might be needed to expand the network if the task is very different from the existing ones or the network capacity is not sufficient when more and more newly coming tasks. And human experts usually find a sub-optimal solution, this encourages us to propose AutoML based network expanding method for lifelong learning. We name it Regularize, Expand, Compress (\pname{}) and summarize the steps in Algorithm~\ref{alg:1}. The details of the network transformations based AutoML for \pname{} are outlined in Algorithm~\ref{alg:Net2Net}. 

We consider net2wider and net2deeper operators~\cite{chen2015net2net} in our controller. The net2wider network transformation function as follows:
\begin{equation}
    \mathcal{\pi}_{wider}(j) = \left\{
             \begin{array}{lr}
             j & j \leq O_l, \\
             random\ sample\ from\ \{1, ..., O_l\} & j > O_l,
             \end{array}
\right.
\label{eq:net2wider}
\end{equation}
where $O_l$ represents the outputs of the original layer $l$. And the net2deeper network transformation function is
\begin{equation}
    \gamma(\mathcal{\pi}_{deeper}(j))=\gamma(j) \quad  \forall j.
    \label{eq:net2deeper}
\end{equation} 
where the constraint $\gamma$ holds for the rectified linear activation. We learn a meta-controller to generate network transformation actions (Eq.~\ref{eq:net2wider} and Eq.~\ref{eq:net2deeper}) when given the initial network architecture. Specifically, we use an encoder network~\cite{cai2018efficient}, which is implemented with an input embedding layer and a bidirectional recurrent neural network~\cite{schuster1997bidirectional}, to learn a low-dimensional representation of the initial network and be embedded into different operators to generate different network transformation actions. Besides, we use a shared sigmoid classifier to make the Net2Wider decision according to the hidden state of the layer learned by the bidirectional encoder network~\cite{cai2018efficient} and the wider network can be further combined with a Net2Deeper operator. 

We then integrate MWC (Eq.~\ref{eq:MWC}) into above AutoML system for lifelong learning. After we learning the network $\theta^{t-1}$ on the data $\mathbf{D}^{t-1}$, we will automatically search the best child network $\theta^t$ by Net2wider and Net2Deeper operators when it is necessary to expand the network while keeping the model performance on task $t-1$ based on Eq.~\ref{eq:MWC}. If the controller decides to expand the network, the newly added layer will not have the previous tasks' Fisher Information. We consider the newly added layer as a new task-specific layer, $l_1$ regularization is adopted to promote sparsity in the new weight so that each neuron only connected with few neurons in the layer below and this will efficiently learn the best representation for the new task while reducing the computation overheads. The modified MWC in network expanding scenario as follows: 
\begin{equation}
\begin{aligned}
   \min_{\theta^t}\quad & \mathcal{F}_t(\theta^t) + \frac{\lambda}{2}\sum_{\substack{i\neq deeper\\i\neq wider}}\mathbb{F}_{i}(\theta^{t}_i - \bar{\theta}^{t-1}_i)^2 \\&+\lambda_2||[\theta^{t};\theta^{t-1}]||_{2, 1}+\lambda_3||\theta^t_{\substack{i=deeper\\i= wider}}||_1,
\end{aligned}
\label{eq:modified-MWC}
\end{equation}
where the subscript ${deeper}$ and ${wider}$ refer to the newly added layer in task $t$.

After the controller generates the child network, the child network will achieve an accuracy $A_{val}$ on the validation set of task $t$ and this will be used as the reward signal $R^t$ to update the controller. We maximize the expected reward to find the optimal child network. The empirical approximation of our AutoML \textsc{reinforce} rule~\cite{sutton2000policy} as follows:
\begin{equation}
\label{eq:autoML}
    \frac{1}{m}\sum_{i = 1}^m\sum_{s=1}^{S}\bigtriangledown_C\log P(a_s|a_1, \cdots, a_{s-1}; C)R ^t_i,
\end{equation}
where $m$ is the number of children networks that the controller $C$ samples in one batch and $a_s$ and $g_s$ represents the action and state of predicting $s$-th hyperparameter to design a child network architecture, respectively. $T$ is the transition function in Alg.~\ref{alg:Net2Net}. Since $R^t$ is non-differentiable, we use policy gradient to update the controller. We use a non-linear transformation $tan(A_{val} \times \pi / 2)$ on validation set of task $t$ as done in~\cite{cai2018efficient} and use the transformed value as the reward. We also use an exponential moving average of previous rewards with a decay of 0.95 to reduce the variance. To balance the old task and new task knowledge, we set maximum expanding layers are 2 and 3 on net2wider and net2deeper operators, respectively. 

If the network keeps expanding as more and more tasks will be given, the model will suffer the inefficient problem and have extra memory cost. Thus, the model compression technique is needed to reduce the memory cost and receive a nonexpansive model. Here, we use soft-label (the logits) as knowledge distillation (KD)~\cite{hinton2015distilling} instead of the hard labels to train the student model. We follow Ba et al.~\cite{ba2014deep} that the student model is trained to minimize the mean of the $l_2$ loss on the training data $\{\mathbf{x}^t_i, z^t_i\}_{i=1}^{N^t}$, where $z_i^t$ is the logits of the child model $\theta^t$ $i$-th training sample. We compress the $\theta^t$ to the same size model as $\theta^1$ by KD loss below:
\begin{equation}
\label{eq:kd}
    \min_{\theta^{t}_c}\mathcal{F}_{kd}(f(\mathbf{x}^t; \theta^{t}_c), \mathbf{z}^t) = \frac{1}{N^t}\sum_{i}||f(\mathbf{x}_i^t; \theta^t_c) - z^t_i||^2_2,
\end{equation}
where $\theta^t_c$ is the weights of the student network and $f(\mathbf{x}^t_i; \theta^t_c)$ is the prediction of task $t$ $i$-th training sample. 
The final student network $\theta^t_c$ is trained to convergence with hard and soft labels by the following loss function:
\begin{equation}
    \min_{\theta^{t}_c}\mathcal{F}(f(\mathbf{x}^t; \theta^t_c), \mathbf{y}^t) + \mathcal{F}_{kd}(f(\mathbf{x}^t; \theta^t_c), \mathbf{z}^t),
    \label{eq:comploss}
\end{equation}
where $\mathcal{F}$ is the loss function (cross-entropy in this work) for training with ground truth $\mathbf{y}^t$ of task $t$. 

\begin{algorithm}[t]
\caption{Automatically Network Transformation}
\label{alg:Net2Net}
\SetKwData{Left}{left}\SetKwData{This}{this}\SetKwData{Up}{up}
\SetKwFunction{Union}{Union}\SetKwFunction{FindCompress}{FindCompress}
\SetKwInOut{Input}{Input}\SetKwInOut{Output}{Output}
\Input{Dataset $\mathbf{D}_t$, $\theta^{t-1}$}
\Output{ The best expended model $\theta^t$}
\Begin{
   \For{ $i=1 \rightarrow m$ }{
       \For{ $s=1 \rightarrow S$ }{
           $a_s \leftarrow \pi_{deeper}(g_{s-1}; \theta^{t-1}_{deeper})$ or $\pi_{wider}(g_{s-1}; \theta_{wider}^{t-1})$\\
           $g_s \leftarrow T(g_{s-1}, a_s)$\\
           $\theta^t \leftarrow \theta^t_{newLayer}$
        }
       $R_i \leftarrow  tanh(A^t_i(g_{S}) \times \pi/2)$\\
       $\theta^t_{i} \leftarrow \bigtriangledown_{\theta^t_{i-1}}J(\theta^t_{i-1})$
    }
}
\end{algorithm}

\begin{figure*}[t]
\centering
\vspace{-1em}
\includegraphics[height=4.5cm, width=5.75cm]{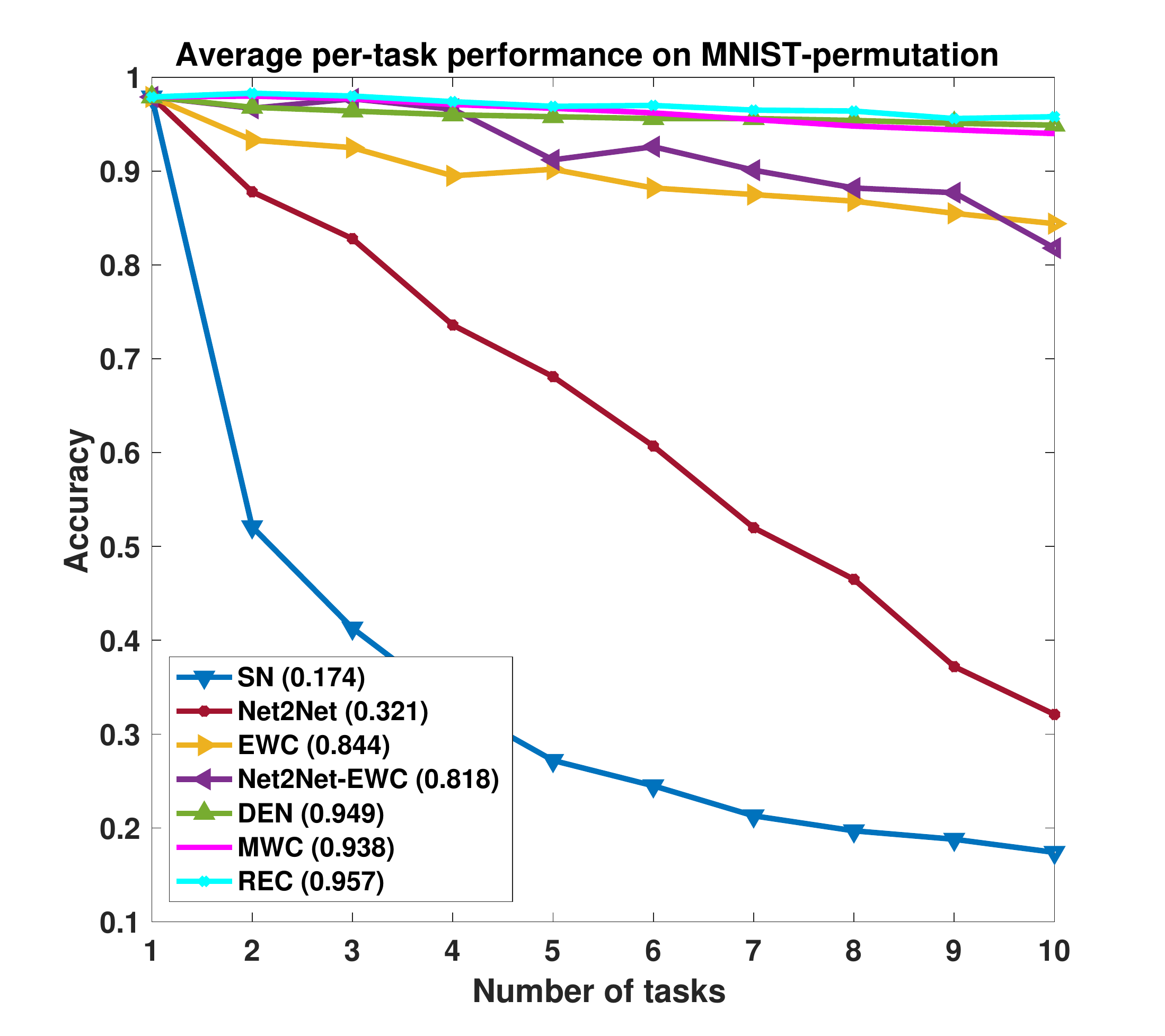}
\includegraphics[height=4.5cm, width=5.75cm]{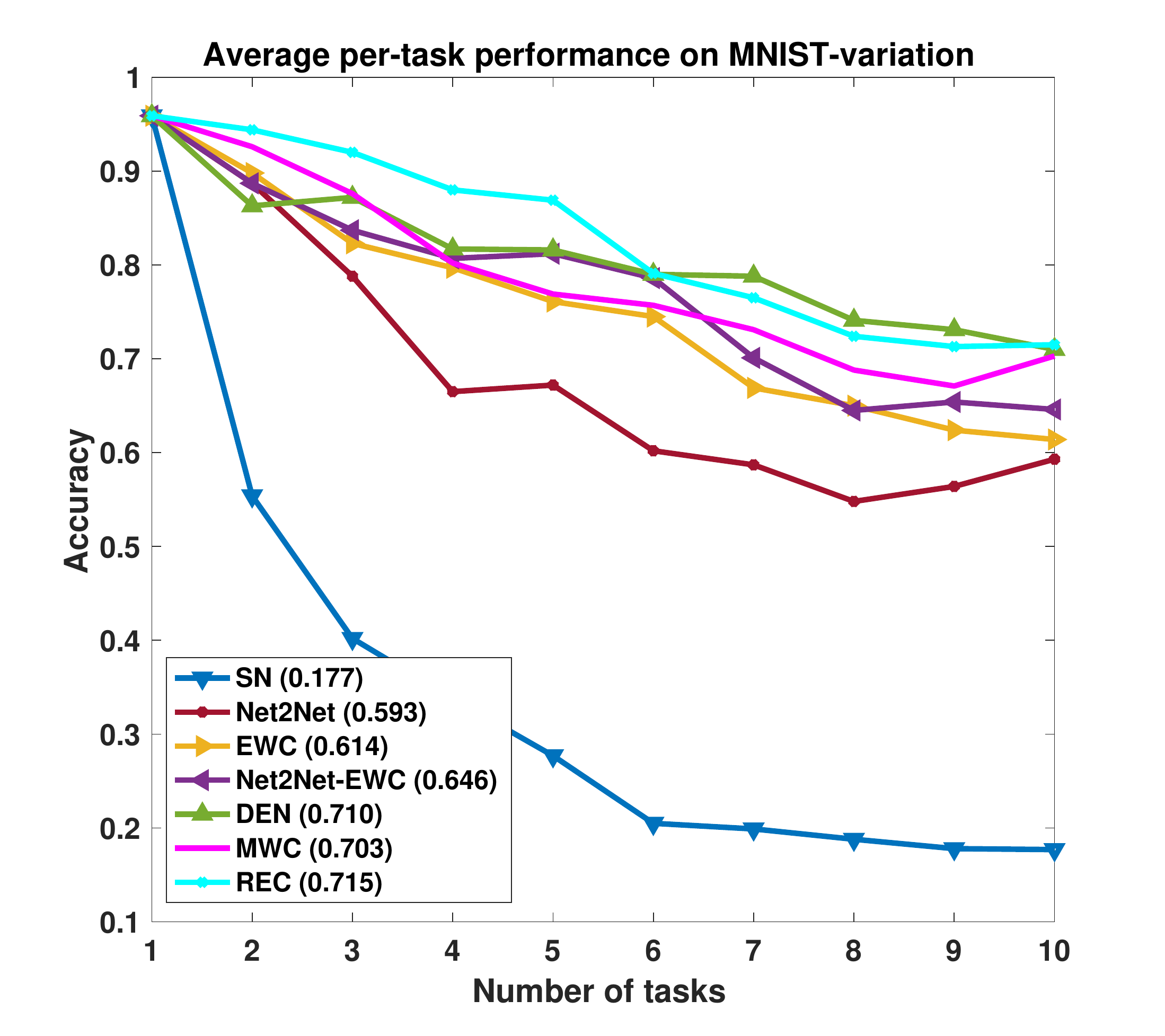}
\includegraphics[height=4.5cm, width=5.75cm]{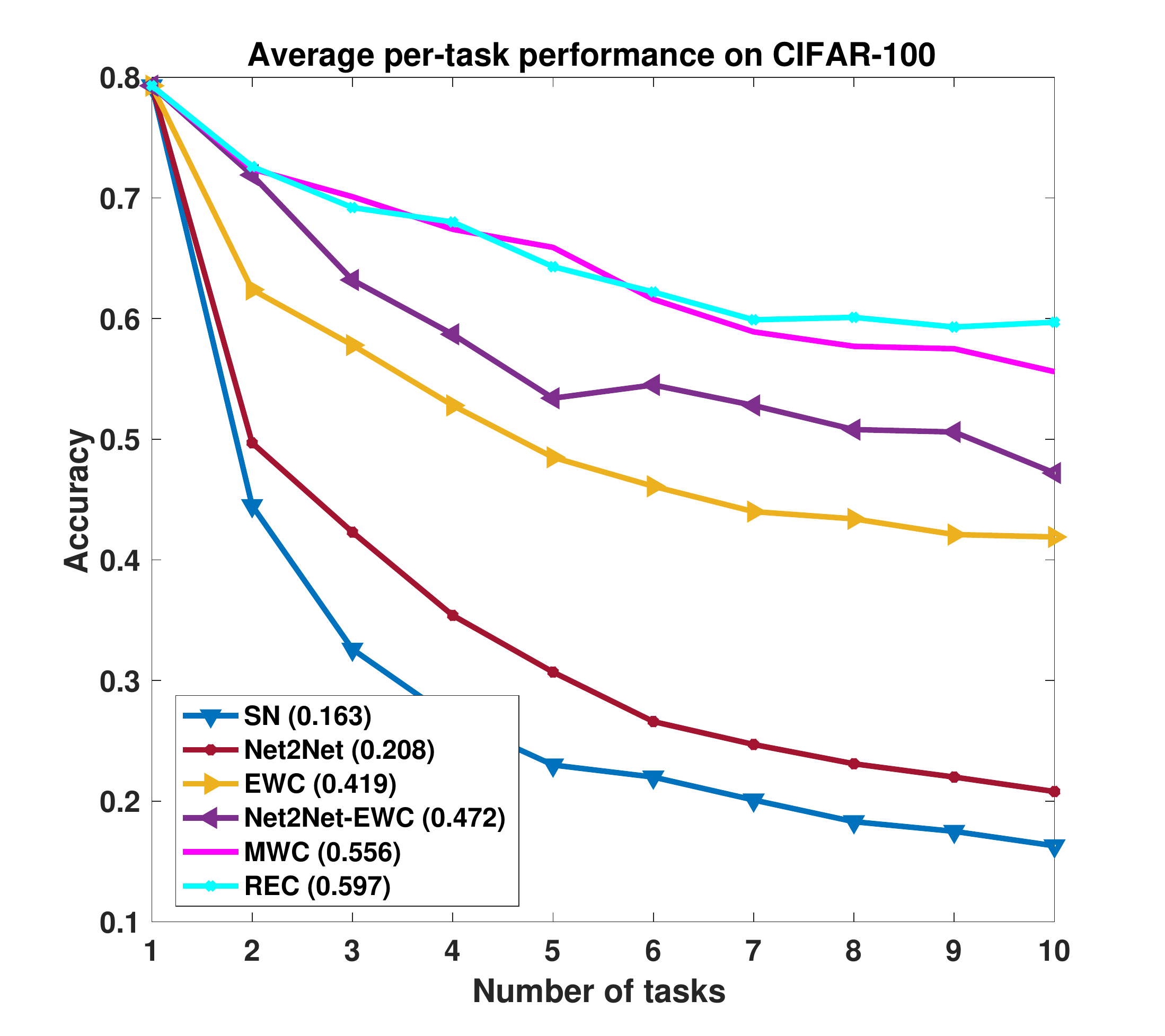}
\caption{The experimental results of continual training on MNIST-permutation, MNIST-variation and CIFAR-100 datasets. We report the average per-task performance (Accuracy) of the models over $T=10$ task. The numbers in the legend represent average per-task performance after the model has finished learning task $t$.}
\label{fig:results-4dataset}
\end{figure*}

\section{Experiments}
\begin{figure*}[t]
\centering
\includegraphics[height=4.5cm, width=5.75cm]{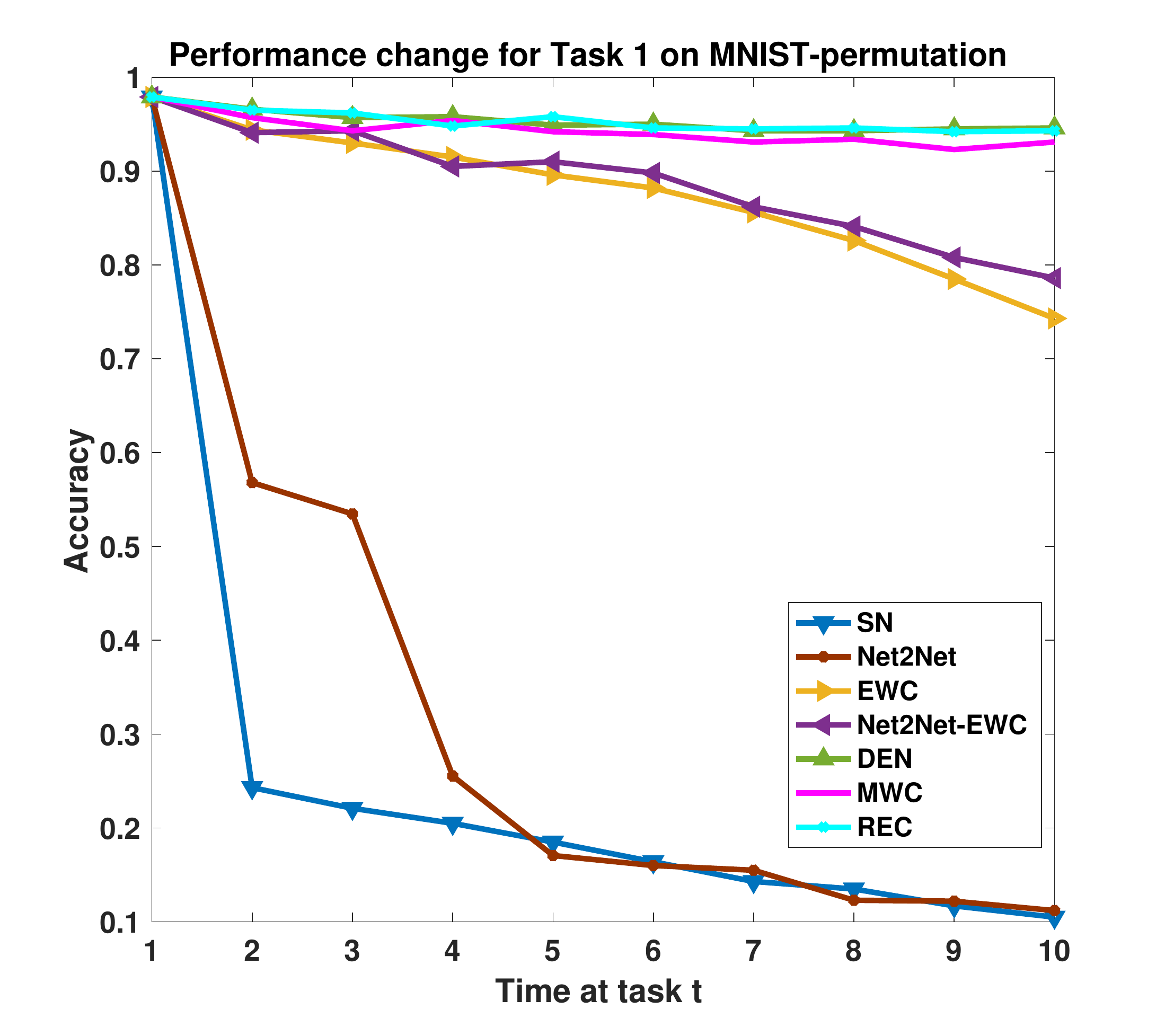}
\includegraphics[height=4.5cm, width=5.75cm]{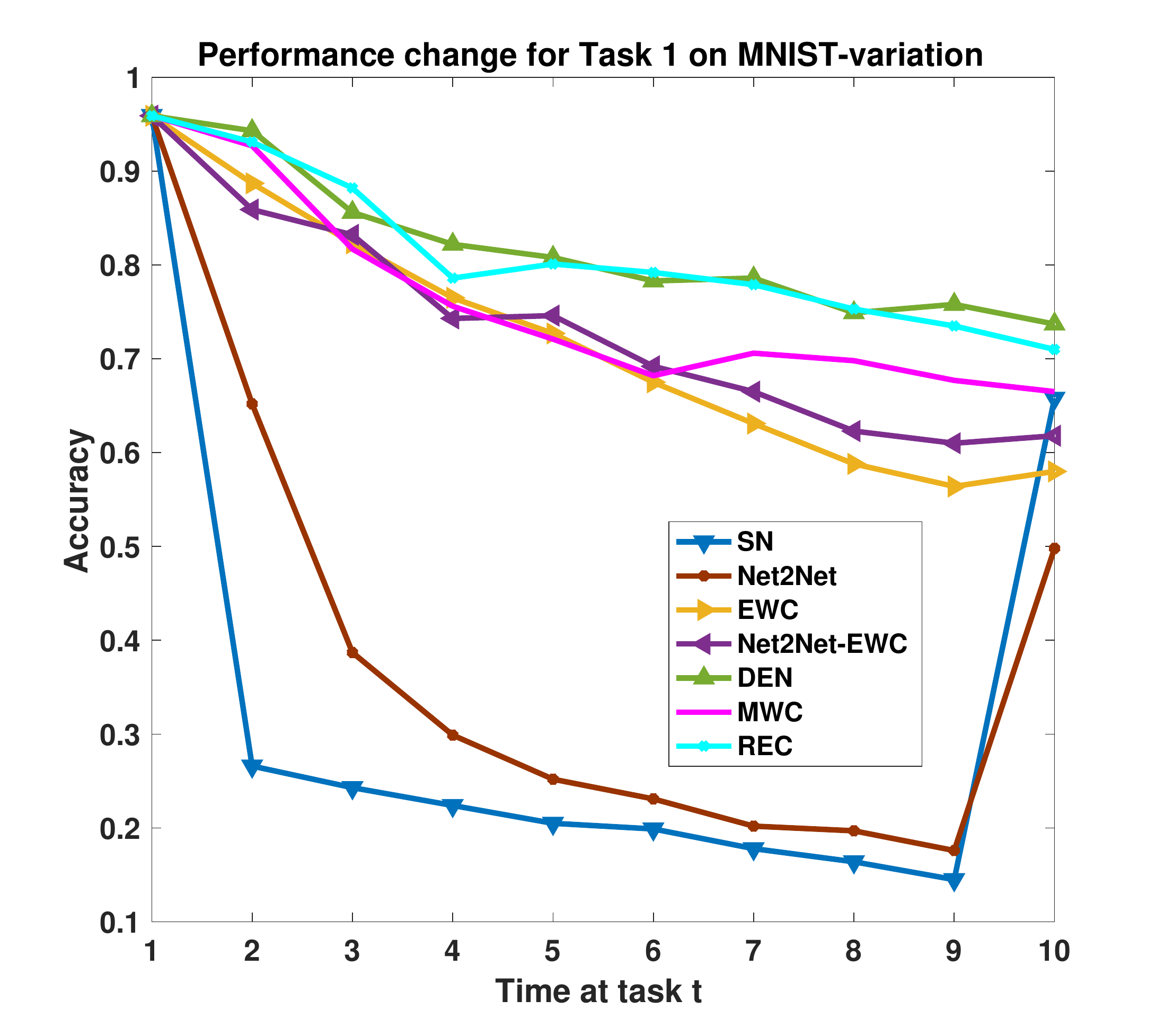}
\includegraphics[height=4.5cm, width=5.75cm]{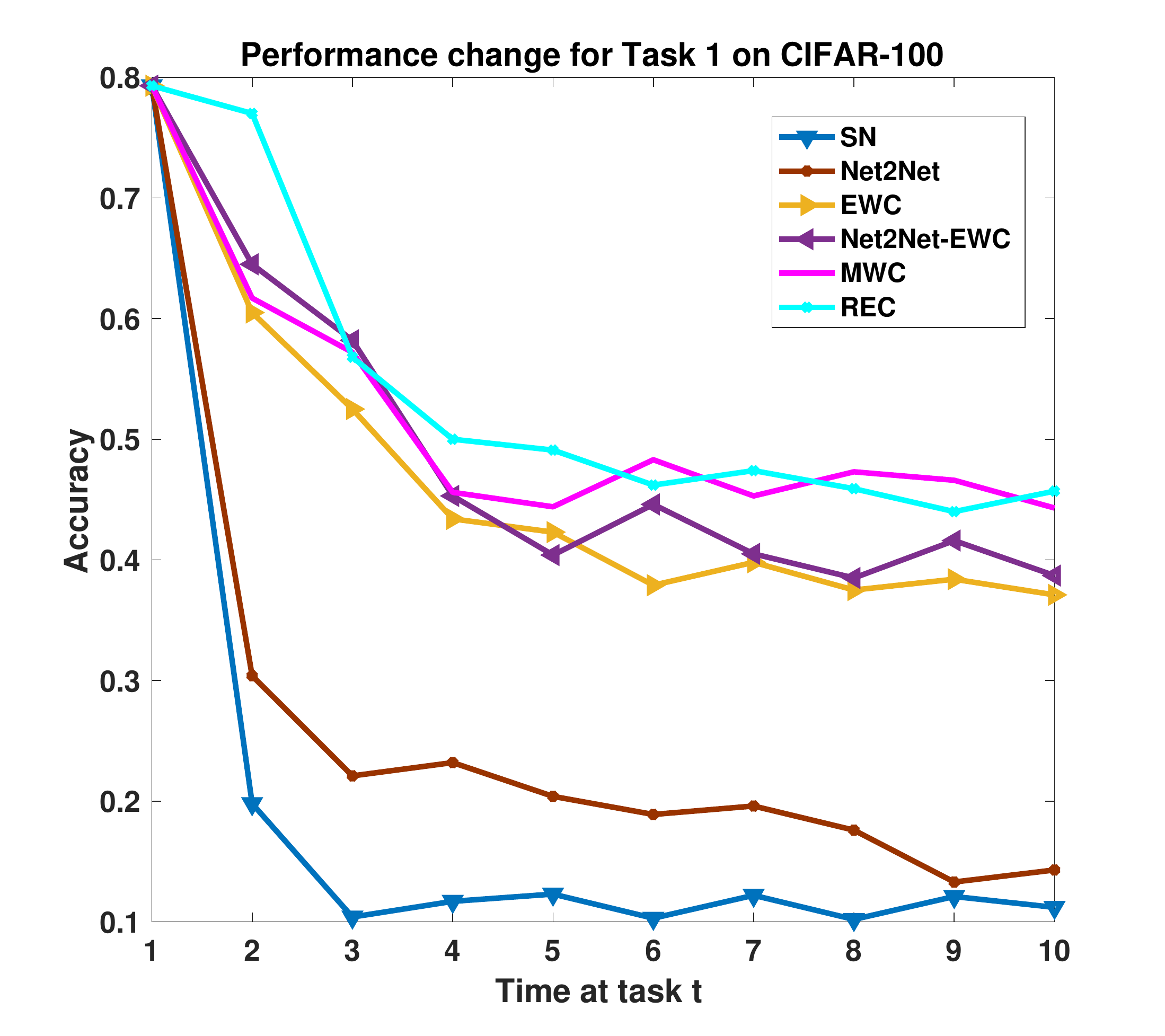}
\caption{Forgetting experiment for task 1 on MNIST-permutation, MNIST-variation and CIFAR-100 datasets. We report the accuracy of different models on task $t = 1$ at each training stage to see how the model performance changes over time for all datasets.}
\label{fig:task1-4dataset}
\end{figure*}


\subsection{Experimental Settings}
\textbf{Datasets.} We evaluate our algorithm on most commonly used datasets for lifelong learning. We list them as follows:

\textbf{-- MNIST-permutation}: MNIST~\cite{lecun1998mnist} is used as the most common datasets among all lifelong learning works, which consists of ten handwritten digits classes with 60,000/10,000 training and testing examples. One way to create the datasets for multiple tasks is randomly permuting the pixels by a fixed permutation~\cite{kirkpatrick2017overcoming} so that the input distribution for each task is unrelated. 

\textbf{-- MNIST-Variation}: MNIST-variation~\cite{lecun1998mnist} dataset rotates the MNIST dataset by a fixed angle between 0 to 180 degrees for each different task. We use $180/T$ as the fixed angle to create $T$ tasks.

\textbf{-- CIFAR-100}: CIFAR-100~\cite{krizhevsky2009learning} dataset contains 60,000 $32\times32$ color images in 100 object classes. Each class has 500/100 images for training and testing. We consider each task with a set of classes, it contains $100/T$ classes when there are $T$ tasks. Different from MNIST-permutation dataset, the input distributions are similar for all tasks but the output distributions for each task are different. 

\textbf{-- CUB-200}: CUB-200~\cite{WahCUB_200_2011} is a fine-grained image classification benchmark, we use CUB-200-2011 version in this work. It contains 11,788 images of 200 types of birds with 5,994/5,794 for training and testing. Each image has detailed annotations and a bounding box. We crop the bounding boxes from the original images and resize them to $224\times224$. We use the same way to create multiple tasks as CIFAR-100 dataset. 

For the first three datasets, we choose $T=10$ tasks. Since the fine-grained CUB-200 dataset is more challenging than others, we set $T=4$ tasks to show better comparisons on lifelong learning. For all datasets, we use 0.1 ratios to split validation set and the model observes the tasks in sequence. We generate multiple tasks for each dataset first and all comparison methods then use the same task order and the same categories within the task for fair comparisons.  

\textbf{Base network settings.} For two MNIST datasets, we use a two-layer fully-connected neural network of 100-100 units with ReLU activations as our initial network. For CIFAR-100 dataset, we use a modified version of AlexNet~\cite{krizhevsky2012imagenet} which has five convolutional layers (64-128-256-256-128 depth with $5\times5$ filter size), and three fully-connected layers (384-192-100 neurons at each layer) and the standard data augmentation is used in this dataset. For CUB-200 dataset, we use a pre-trained VGG-16~\cite{simonyan2014very} model from ImageNet~\cite{deng2009imagenet} and fine-tune it on the CUB-200 data for better initialization. We follow the setting of Liu et al.~\cite{liu2018rotate}, which adds a global pooling layer after the final convolutional layer of the VGG-16. The fully-connected layers are changed to 512-512 and the size of the output layer is the number of classes in each task. All models and algorithms are implemented using Tensorflow~\cite{abadi2016tensorflow} library.

\textbf{Comparison methods.} We compare our algorithm with six other methods: 1) SN: A single network trained across all tasks. 2) Net2Net~\cite{chen2015net2net}: Network expanding by Net2Net~\cite{chen2015net2net} on new task. 3) EWC~\cite{kirkpatrick2017overcoming}: A deep network trained with elastic weight consolidation. 4) Net2Net-EWC: Network expanding by Net2Net~\cite{chen2015net2net} with elastic weight consolidation~\cite{kirkpatrick2017overcoming} when learning new task. 5) DEN~\cite{yoon2018lifelong}: Dynamically expandable network. 6) REWC~\cite{liu2018rotate}: Rotate Elastic Weight Consolidation. 7) MWC: A deep network trained with multi-task weight consolidation. 8)\pname{}: Regularize, Expand and Compress.

\textbf{Hyperparameter settings.} All hyper-parameters in MWC are optimized using a grid-search and the best results for each model are reported. For two MNIST datasets, the SGD optimizer is used with a learning rate of $0.001$ and we set batch size of 256 with 8 epochs, $\lambda_1 = 2$, $\lambda_2 = 0.0001$ and $\lambda_3 = 0.001$ in all experiments. For CIFAR-100 dataset, we use SGD optimizer with momentum parameter of $0.9$, learning rate of $0.01$, batch size of 128 with 20 epochs, $\lambda_1 = 10$, $\lambda_2 = 0.015$ and $\lambda_3 = 0.0001$. For CUB dataset, the Adam optimizer is used with a learning rate of $0.001$, batch size of 32 and 50 epochs, $\lambda_1 = 100$, $\lambda_2 = 0.001$ and $\lambda_3 = 0.005$. For network transformation based AutoML experimental settings, we followed the training details of Cai \etal~\cite{cai2018efficient}.

\subsection{Experimental Results}
We evaluate our methods from both model accuracy and model complexity, where we measure the model size at the end of the training process.

\begin{table}[t]
\centering
\caption{Comparisons of the model size and the average task accuracy after training $10$ tasks of different approaches on MNIST-permutation. $\#W(1)$: the number of parameters of task 1. $\#W(10)$: the number of parameters after training task 10. ACC (10): average per-task accuracy after training task 10.}
 \begin{tabular}{c|c|c|c}\hline
Methods	& $\#$W (1) & $\#$W (10) & ACC (10) \\ \hline
SN &0.01M&0.01M&17.4\%\\
Net2Net&0.01M& 0.02M & 32.1\%\\
EWC&0.01M & 0.01M &84.4\%\\
Net2Net-EWC &0.01M & 0.02M& 81.8\%\\
DEN&0.01M& 0.14M &94.9\%\\
MWC&0.01M &0.01M&93.8\% \\
\pname{}&\textbf{0.01M}&\textbf{0.01M}&\textbf{95.7\%}\\\hline
\end{tabular}
\vspace{-1em}
\label{table:mnist}
\end{table}

\textbf{Comparisons of the model performance.} We report the average per-task accuracy of MNIST-permutation, MNIST-variation and CIFAR-100 datasets when $T=10$ in Fig.~\ref{fig:results-4dataset}. Overall, \pname{} outperforms all comparison methods and overcomes catastrophic forgetting especially on the later tasks (after task 5). We can observe that the regularization based network (EWC, MWC) has worse performance than expandable networks (DEN, \pname{}), which shows that selectively expand networks help improve the performance by a large margin. Specifically, \pname{} performs better than DEN on two MNIST datasets and MWC performs similarly with DEN on MNIST-permutation dataset while using fewer parameters. We also observe that directly apply Net2Net~\cite{chen2015net2net} on lifelong learning does not perform well since it forgets the old tasks' knowledge as finetuning (SN), but adding EWC as the loss function can help enhance the old tasks' performance on Net2Net. \pname{} has better performance than Net2Net-EWC, because we consider the new task-specific parameters and the discriminative common subset between the old tasks and the new one. 

We also evaluate the catastrophic forgetting over time on the earliest task, Fig.~\ref{fig:task1-4dataset} shows the test accuracy of the first task throughout the whole lifelong learning process on MNIST-permutation, MNIST-variation and CIFAR-100 datasets. It shows that our methods (MWC and \pname{}) overcome forgetting on old tasks compared with all other methods on MNIST-permutation and CIFAR-100 datasets. It is worth noting that DEN performs slightly better than our method on task 1 after learning later tasks on MNIST-variation dataset due to they selectively expands network for the new task, it will give a bias towards to the earliest task. Our \pname{} is a nonexpensive network and our overall average per-task performance is better than DEN, which shows that our method has better performance on later learned tasks and achieve a more balanced performance when learning sequential tasks in the temporal dimension comparing with DEN. Besides, we have an interesting founding on MNIST-variation dataset, the SN and Net2Net has irregular performance on task 1 after learning task 10, it is due to the task 10 is the upside-down flipped image of task 1 and such flip gives benefit on some digits such as `1',`0',`8'. And SN and Net2Net forget too much task 1' knowledge after learning task 9, they only can keep the most recently learned task knowledge when they learn task 10 comparing with EWC, MWC and \pname{} and this causes the irregular performance.

\begin{table}[t]
\centering
\vspace{-1em}
\caption{Comparisons of the model size and the average task accuracy after training $10$ tasks of different approaches on CIFAR-100 dataset. $\#W (1)$: the number of parameters of task 1. $\#W(10)$: the number of parameters of the model after training task 10. ACC (10): average per-task accuracy after training task 10.}
\label{table:cifar100}
 \begin{tabular}{c|c|c|c}\hline
Methods	& $\#$W (1) & $\#$W (10) & ACC (10) \\ \hline
SN &4M&4M&16.3\%\\
Net2Net& 4M& 6.3M& 20.8\%\\
EWC&4M &4M&41.9\%\\
Net2Net-EWC &4M& 7.4M& 47.2\%\\
MWC&4M &4M& 55.6\%\\
\pname{}&\textbf{4M}&\textbf{4M}&\textbf{59.7\%}\\\hline
\end{tabular}
\vspace{-1em}
\end{table}

\textbf{Comparisons of the model complexity. }
 Table~\ref{table:mnist} and Table~\ref{table:cifar100} report the comparisons of the model size and the average per-task performance after training $T=10$ tasks of different approaches on MNIST-permutation and CIFAR-100 datasets, respectively. Overall, \pname{} performs similarly or better than all other approaches with smaller model size. We observe that DEN performs better than MWC and worse than \pname{} on MNIST-permutation dataset, but it has 1.4X network expansion comparing with ours. For CIFAR-100 dataset, We compute our AUROC after learning $T=10$ tasks, \pname{} can achieve 0.887 comparing with DEN (0.923), however, our model size is 50\% of DEN's model. Besides, we notice that DEN involves 7 hyperparameters and very sensitive to them, we slightly change one of them from $10^{-3}$ to $10^{-2}$, the result becomes 0.8907 on MNIST-permutation dataset. Our method only has three hyperparameters and it needs much less expert tuning comparing with DEN. Training times is a limitation of the current version of \pname{}, since \pname{} is a reinforcement learning based algorithm, a varies number of trails are needed and this results in more training time than other methods. We will improve the training efficiency of our work in the future. Besides, we did not consider complexity network structures (e.g. ResNet~\cite{he2016deep}, DenseNet~\cite{huang2017densely}), we will extend the current work to more network architectures in the future. 
 
 \begin{figure}[t]
\centering
\vspace{-1em}
\includegraphics[height=4cm]{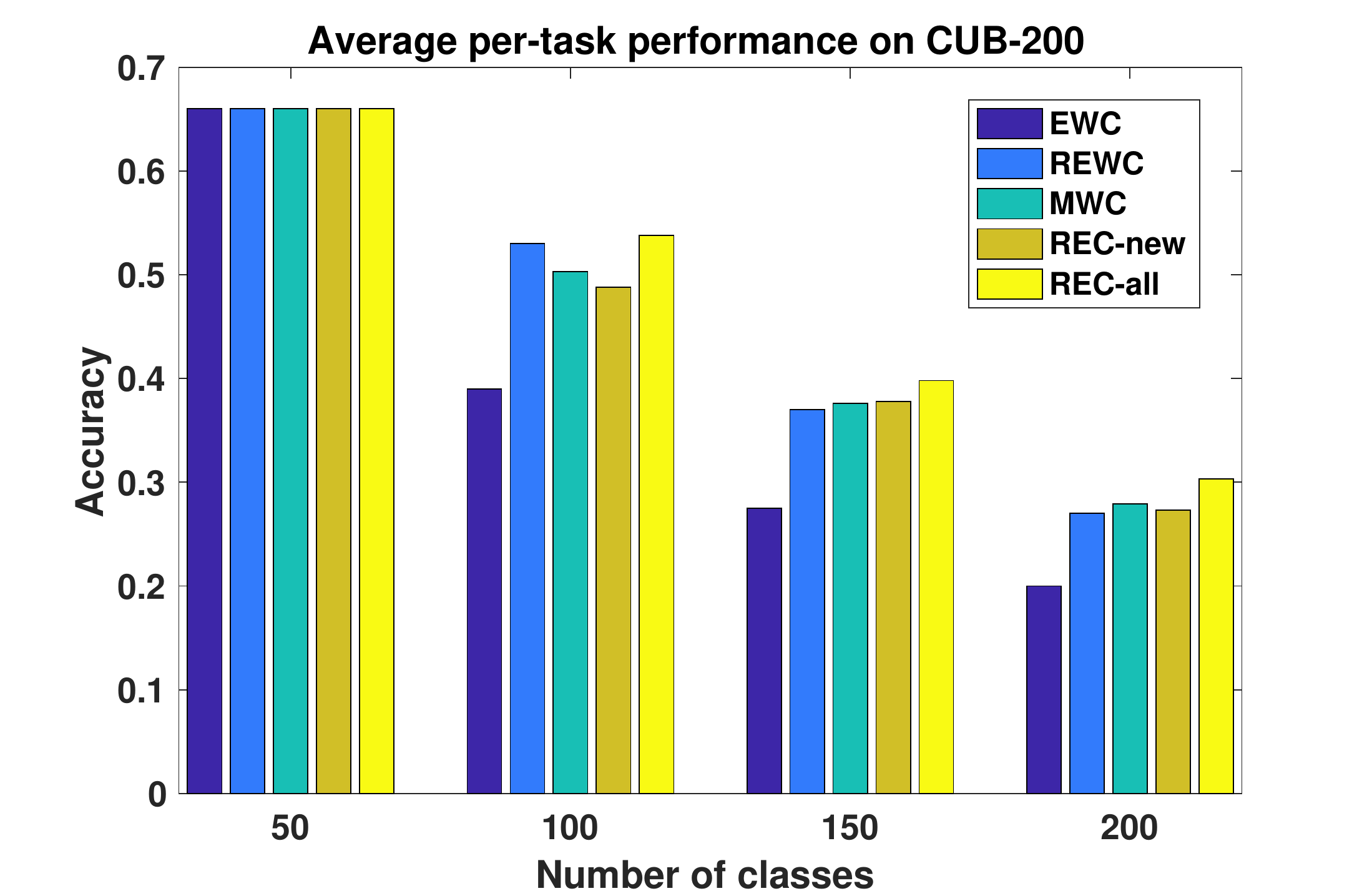}
\caption{Comparison results with EWC and REWC on CUB-200 dataset when $T=4$.}
\label{fig:CUB}
\end{figure}

\textbf{Comparison results on CUB-200 dataset.}
Fig.~\ref{fig:CUB} shows the comparison results when $T=4$ on CUB-200 dataset with EWC~\cite{kirkpatrick2017overcoming} and REWC~\cite{liu2018rotate}. It shows that MWC has comparable results with REWC, MWC has better performance on task 3 and task 4 while has worse performance on task 2. We test REC with only new task validation set (REC-new), which has similar results as MWC on later tasks. This might be caused by using only new task validation set is not sufficient to compute the rewards on a more subtle dataset. We hypothesis the exemplars from old tasks will help improve the nonexpansive AutoML system's performance. Thus, we use the validation sets of all learned tasks to compute the rewards and report the results (REC-all) in Fig.~\ref{fig:CUB}. The results show that exemplars from old tasks help improve the performance of AutoML based algorithm and we will investigate the relationship between the number of exemplars and the performance of \pname{} in our future work.

 \begin{table}[t]
\centering
\caption{Comparison results of average per-task accuracy after training task 10 on MNIST-permutation dataset.}
\label{table:combination}
 \begin{tabular}{c|cccc}\hline
Method	& EWC & EWC+$l_1$ & EWC+$l_{2,1}$ & MWC \\ \hline
ACC(10) &84.4\%&87.7\%&88.5\% &94.0\% \\ \hline
\end{tabular}
\vspace{-1em}
\end{table}

\textbf{Ablation study on each component in MWC.}
We study how the different components used in MWC affect the final performance of lifelong learning. We report the average per-task accuracy after training task 10 on MNIST-permutation of different strategies \textit{EWC}, \textit{EWC with $l_1$-norm only}, \textit{EWC with $l_2$-norm only} and \textit{MWC} in Table~\ref{table:combination}. It shows that $l_{2,1}$-norm has a stronger effect of the performance than $l_1$-norm while our method MWC outperforms the single regularization strategies, which demonstrates the meaningful and useful of our method by studying common weights subset with discriminative new task parameters.

\section{Conclusion and Future Works}
In this work, we develop a multi-task based lifelong learning framework via nonexpansive AutoML (\pname{}). \pname{} is achieved at two stages: continually network expansion and model compression, besides a novel multi-task weight consolidation algorithm is proposed to overcome catastrophic forgetting. We achieved better accuracy and smaller model size than other lifelong learning methods on four datasets. In the future, we plan to reduce the training time of the AutoML based algorithm and explore the need of exemplars for computing the rewards to improve the current work.

{\small
\bibliographystyle{ieee}
\bibliography{egbib}
}

\end{document}